\documentclass[10pt,twocolumn,letterpaper]{article}

\usepackage{wacv}
\usepackage{times}
\usepackage{epsfig}
\usepackage{graphicx}
\usepackage{amsmath}
\usepackage{amssymb}
\usepackage{booktabs}
\usepackage{subcaption}
\usepackage{adjustbox}
\usepackage{hhline}
\def\wacvPaperID{****} % Enter the WACV Paper ID here

\wacvfinalcopy % *** Uncomment this line for the final submission

\ifwacvfinal
\usepackage[breaklinks=true,bookmarks=false]{hyperref}
\else
\usepackage[pagebackref=true,breaklinks=true,colorlinks,bookmarks=false]{hyperref}
\fi
\usepackage{cleveref}
\newcommand{\modulationMetod}{Surface Adaptive Modulation\xspace}
\newcommand{\modulationMetodShort}{SAM\xspace}

\newcommand{\modulationMetodNoiseShort}{V-SAM\xspace}

\newcommand{\paperTitle}[0]{Realistic Full-Body Anonymization with Surface-Guided GANs}
\newcommand{\appendixDetails}[0]{Appendix A\xspace}
\newcommand{\appendixAnonymization}[0]{Appendix B\xspace}
\newcommand{\appendixQuantitative}[0]{Appendix C\xspace}
\newcommand{\appendixQualitative}[0]{Appendix D\xspace}

\newcommand{\authorskip}{\hspace{10mm}}
\newcommand{\paperAuthor}[0]{
    H{\aa}kon Hukkel{\aa}s \authorskip Morten Smebye \authorskip Rudolf Mester \authorskip Frank Lindseth \\
Norwegian University of Science and Technology \\
{\tt\small hakon.hukkelas@ntnu.no}
}
\newcommand{\appendixref}[1]{\hyperref[#1]{Appendix \ref*{#1}}}

\begin{document}
%%%%%%%%% TITLE

%%%%%%%%% TITLE
\title{\paperTitle}

\author{\paperAuthor}
\twocolumn[{%
\renewcommand\twocolumn[1][]{#1}%
\maketitle
\begin{center}
    \centering
    \captionsetup{type=figure}
    \includegraphics[width=\linewidth]{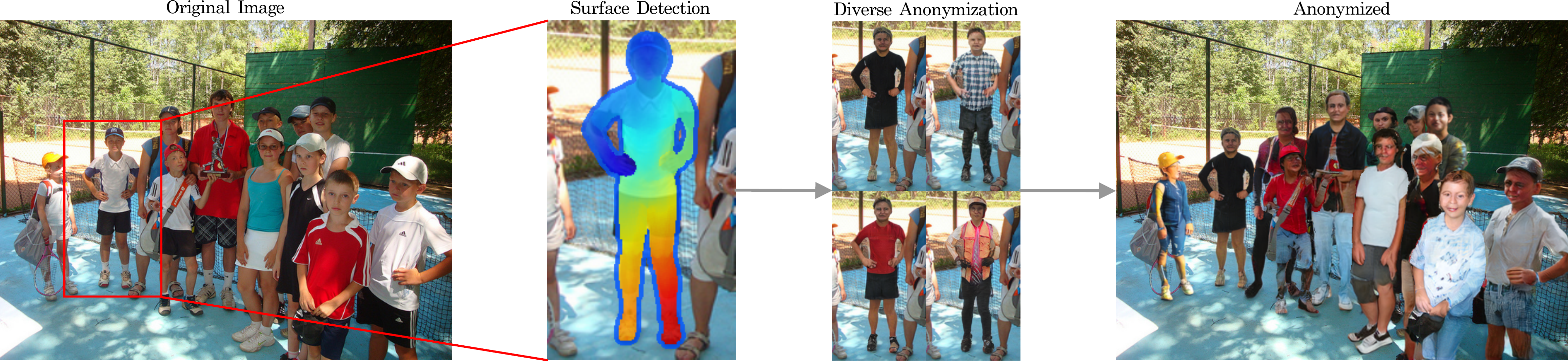}
    \captionof{figure}{Our model performs in-the-wild anonymization by first detecting pixel-to-surface correspondences with CSE \cite{neverova2020cse}, then Surface-Guided GAN individually anonymizes each person.
  Original image from COCO \cite{lin2014microsoft}.}
    \label{fig:task_figure}
\end{center}%
}]

\begin{abstract}
Recent work on image anonymization has shown that generative adversarial networks (GANs) can generate near-photorealistic faces to anonymize individuals.
However, scaling up these networks to the entire human body has remained a challenging and yet unsolved task.
We propose a new anonymization method that generates realistic humans for in-the-wild images.
A key part of our design is to guide adversarial nets by dense pixel-to-surface correspondences between an image and a canonical 3D surface.
We introduce Variational Surface-Adaptive Modulation (V-SAM) that embeds surface information throughout the generator.
Combining this with our novel discriminator surface supervision loss, the generator can synthesize high quality humans with diverse appearances in complex and varying scenes.
We demonstrate that surface guidance significantly improves image quality and diversity of samples, yielding a highly practical generator.
Finally, we show that our method preserves data usability without infringing privacy when collecting image datasets for training computer vision models.
Source code and appendix is available at: \href{https://github.com/hukkelas/full_body_anonymization}{github.com/hukkelas/full\_body\_anonymization}
\end{abstract}

\section{Introduction}
Privacy regulations constitute a significant obstacle against using image data taken in public for training computer vision algorithms.
Recent work reflects that generative adversarial networks (GANs) \cite{hukkelaas2019DeepPrivacy,maximov2020ciagan,sun2018natural} can realistically anonymize faces, where the anonymized datasets perform similarly to the original for future computer vision development.
However, these methods \cite{hukkelaas2019DeepPrivacy,maximov2020ciagan,sun2018natural} focus solely on face anonymization, leaving several primary  identifiers(\eg ears \cite{Jain2008handbookOfBiometrics}) and soft identifiers (\eg gender) on the human body untouched.

Generative adversarial networks are great at synthesizing high-resolution images in many domains, including humans \cite{karras2019style}.
Despite this success, previous work on full-body generative modeling focuses on simplified tasks, such as motion transfer \cite{chan2019everybody}, pose transfer \cite{balakrishnan2018synthesizing,li2019dense}, garment swapping \cite{han2018viton}, or rendering a body with known 3D structure of the scene \cite{weng2020vid2actor}.
These methods do not directly apply to in-the-wild anonymization, as they do not handle variations in the background. 
As far as we know, our work is the first to address the task of synthesizing humans into in-the-wild images without simplifying the task (\eg having a source texture to transfer, known 3D structure of the scene, or assuming a static background) \footnote{Although, we note that CIAGAN \cite{maximov2020ciagan} ablates their method for low-resolution human synthesis.}.

Our contributions address the unexplored and challenging task of full-body anonymization for in-the-wild images.
Our goal is to ensure the privacy of the anonymized individual; thus, we pose the anonymization task as an image inpainting problem.
Modeling anonymization as image inpainting has stronger privacy guarantees than previous human synthesis methods, which rely on a source body texture or the original identity.

In this work, we propose \emph{Surface-guided GANs} that utilize Continuous Surface Embeddings (CSE) \cite{neverova2020cse} to guide the generator with pixel-to-surface correspondences.
The compact, high-fidelity, and continuous representation of CSE excels for synthesizing human figures, as it allows for simple modeling choices without compromising fine-grained details.
We show that surface guidance  significantly improves image quality, whereas current state-of-the-art GANs struggle with generating human figures without it.

We summarize our contributions into three points.

First, to efficiently utilize the powerful CSE representation, we propose \textit{Variational Surface Adaptive Modulation} \textit{(\modulationMetodNoiseShort)}.
\modulationMetodNoiseShort projects the input latent space of the generator to an intermediate surface-adaptive latent space.
This allows the generator to directly map the latent factors of variations to relevant surface locations (\eg relate "red shirt" to the upper body independent of its spatial position), resulting in a latent space disentangled from the spatial image.
The explicit disentangled representation is unique to \modulationMetodNoiseShort, which significantly improves latent disentanglement and image fidelity compared to previous spatially-invariant \cite{karras2019style,zhao2021comodGAN} and spatial-adaptive modulation \cite{Park_2019}.

Secondly, we propose \emph{Discriminator Surface Supervision} that incentivizes the discriminator to learn pixel-to-surface correspondences.
The surface awareness of the discriminator provides higher-fidelity feedback to the generator, which significantly improves image quality.
In fact, we find that the surface-aware feedback from the discriminator is a key factor to the powerful representation learned by \modulationMetodNoiseShort , where similar semantic-based supervision \cite{schonfeld2020oasis} yields suboptimal results.

Thirdly, we present a novel full-body anonymization framework that produces close-to-photorealistic images.
We demonstrate that surface-guided anonymization significantly improves upon traditional methods (\eg pixelation) in terms of data usability and privacy.
For example, pixelation degrades the person average precision by 14.4 for Mask R-CNN \cite{he2017mask} instance segmentation.
In contrast, surface-guided anonymization yields only a 2.8 degradation.

\begin{figure*}[t]
    \centering
    \includegraphics[width=\linewidth]{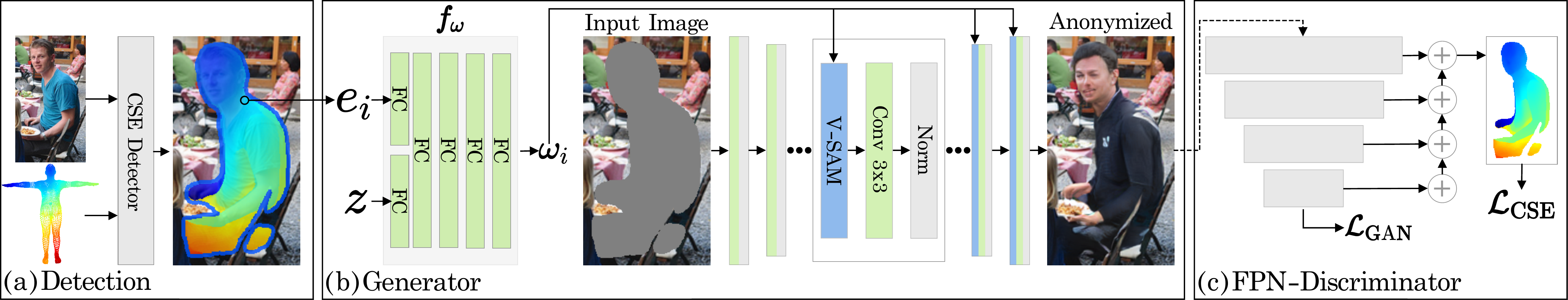}
    \caption{
        \textbf{(a)} A CSE-detector \cite{neverova2020cse} predicts pixel-to-surface correspondences represented as a continuous positional embedding $e_i$.
        For simplicity, we show the pipeline with a single person, but multi-person detection is done by cropping each person (see \Cref{fig:task_figure}).
        \textbf{(b)} The mapping network ($f_\omega$) transform surface locations and the latent variable ($z \sim \mathcal{N}(0, 1)$) into an intermediate surface-adaptive latent space ($\omega_i$) (\cref{sec:surface_adaptive_modulation}, \cref{sec:variational_surface_adaptive_modulation}).
        Then, $w_i$ controls the generator with pixel-wise modulation and normalization after each convolution.
        \textbf{(c)} Our FPN-discriminator predicts the surface embedding and optimizes a surface-regression loss ($\mathcal{L}_{\text{CSE}}$, \cref{sec:surface_regression}) along with the adversarial loss ($\mathcal{L}_{\text{GAN}}$).
        }
    \label{fig:synthesis_architecture}
\end{figure*}

\section{Related Work}

\paragraph{Anonymization of Images}
Naive anonymization methods that apply simple image distortions (\eg blurring) are known to be inadequate for removing privacy-sensitive information \cite{gross2006model,neustaedter2006blur}, and severely distorts the data. 
Recent work reflects that deep generative models can realistically anonymize faces by inpainting \cite{balaji2021temporally,hukkelaas2019DeepPrivacy,maximov2020ciagan,sun2018natural,sun2018hybrid} or transforming the original image \cite{gafni2019live}.
These methods demonstrate that retaining the original data distribution is important for future computer vision development (\eg evaluation of face detection \cite{hukkelaas2019DeepPrivacy}).
However, prior work focuses on face anonymization, leaving several primary and secondary identifiers untouched.
Some methods anonymize the entire body \cite{brkic2017know,maximov2020ciagan}, but these methods are limited to low-resolution images \cite{maximov2020ciagan} or generate  images with visual artifacts \cite{brkic2017know}.

\paragraph{Conditional Image Synthesis}
Current state-of-the-art for conditional image synthesis generates highly realistic images in many domains, such as image-to-image translation \cite{Isola_2017,schonfeld2020oasis}.
An emerging approach is to introduce conditional information to the generator via \emph{adaptive modulation} (also known as adaptive normalization \cite{Huang_2017}).
This is known to be effective for unconditional synthesis \cite{karras2019style}, semantic synthesis \cite{Park_2019}, and style transfer \cite{Huang_2017}.
Adaptive modulation conditions the generator by layer-wise shifting and scaling feature maps of the generator, where the shifting and scaling parameters are adaptive with respect to the condition.
In contrast to prior semantic-modulation methods \cite{Park_2019,tan2021INADE,Tan_2021CLADE}, \modulationMetodNoiseShort conditions the modulation parameters on dense surface information and generates global modulation parameters instead of independent layer-wise parameters.
Conditional modulation is adapted for human synthesis, where prior methods adapt spatially-invariant \cite{men2020controllable,sarkar2021style}, or spatially-variant modulation \cite{albahar2021pose,zhang2021pise}.
However, these methods are conditioned on a source appearance, yielding softer privacy guarantees compared to \modulationMetodNoiseShort.

\paragraph{Human Synthesis}
Prior work for person image generation often focus on resynthesizing humans with user-guided input, such as rendering persons in novel poses \cite{balakrishnan2018synthesizing,li2019dense}, with different garments \cite{han2018viton}, or with a new motion \cite{chan2019everybody}.
Recent work \cite{chaudhuri2021semi,grigorev2019coordinate,lazova2019360,neverova2018dense,sarkar2020neural} employ dense pixel-to-surface correspondences in the form of DensePose UV-maps \cite{Guler_2018}.
These methods "fill in" UV texture maps, then render the person in new camera views \cite{chaudhuri2021semi} or poses \cite{grigorev2019coordinate,lazova2019360,neverova2018dense,sarkar2020neural}.
In contrast, CSE is a much more compact representation, and the continuous representation eases modeling complexity (\eg downsampling of DensePose is not straightforward) and removes the need to handle borders.
In other cases, the aim is to reconstruct the 3D surface and texture \cite{natsume2019siclope,saito2019pifu,weng2020vid2actor}, which can be rendered to the scene given a camera view \cite{weng2020vid2actor}.
A limited amount of work focuses on human synthesis without a source image, where Ma \etal \cite{ma2018disentangled} maps background, pose, and person style into Gaussian variables, enabling synthesis of novel persons.
None of these methods are directly applicable for human anonymization, as they require information about a source identity or the camera position to render the person.
Additionally, none of them account for modeling background variations in the scene, which is the challenge of in-the-wild anonymization.
\section{Method}
\label{sec:method_generative}

We describe the anonymization task as an inpainting task.
The objective of the generator is to inpaint the missing regions in the image $I \odot M$, where $M_i=0$ for missing pixels and 1 otherwise.
For each missing pixel, the surface embedding $e_i \in \mathbb{R}^{16}$ (the output of a CSE-detector \cite{neverova2020cse}) represents the position of pixel $i$ on a canonical 3D surface $S\;$ (\ie the position on a "T-shaped" human body).
The surface $S$ is discretized with $27K$ vertices, where each vertex has a positional embedding $e_k$ obtained from the CSE-detector \cite{neverova2020cse}.
From this, pixel-to-vertex correspondences are found from euclidean nearest neighbor search between $e_i$ and $e_k$ 
\footnote{
    Finding pixel-to-vertex correspondence is not strictly necessary.
    However, replacing the regressed embedding $e_i$ with the nearest $e_k$ prohibits the generator from directly observing embeddings regressed from the original image.
    This can mitigate identity leaking through CSE-embeddings.
}.
\Cref{fig:synthesis_architecture} shows the overall architecture.

\subsection{\modulationMetod}
\label{sec:surface_adaptive_modulation}
Inspired by the effectiveness of semantic-adaptive modulation \cite{Park_2019}, we introduce \emph{\modulationMetod} (\modulationMetodShort).
\modulationMetodShort normalizes and modulates convolutional feature maps with respect to dense pixel-to-surface correspondences between the image and a fixed 3D surface.
Given the continuous positional embedding $e_i$, a non-linear mapping $f_\omega$ transforms $e_i$ to an intermediate surface-adaptive representation $\omega_i$;
\begin{equation}
    \omega_i = \begin{cases}
        f_\omega(e_i) & \text{if } M_i =0 \\
        \omega_M & \text{otherwise}
    \end{cases},
    \label{eq:f_w}
\end{equation}
where $\omega_i \in \mathbb{R}^D$, and $\omega_M \in \mathbb{R}^D$ is a pixel-independent learned parameter for all pixels that do not correspond to the surface ($D=512$ for all experiments).
Given $\omega_i$, a learned affine operation transforms $\omega_i$ to layer-wise "styles" $\gamma_i^\ell$ (we use the word "style" following prior work \cite{Huang_2017,karras2019style}) to scale the feature map $x^\ell$;
\begin{equation}
    \text{SAM}(x^\ell_i, \gamma_i^\ell) = \gamma_i^\ell \cdot x^\ell_i, 
\end{equation}
where each pixel $i$ is modulated by  $\gamma_i^\ell$ independently.
Note that we follow StyleGAN2 design \cite{karras2019analyzing}, with modulation before convolution and normalization after.

\begin{figure}[t]
    \centering
    \begin{subfigure}[t]{0.166666667\linewidth}
        \includegraphics[width=\linewidth]{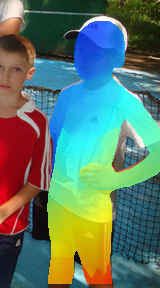}
    \end{subfigure}%%
    \begin{subfigure}[t]{0.166666667\linewidth}
        \includegraphics[width=\linewidth]{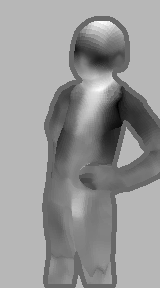}
        \caption{$n=0$}
    \end{subfigure}%%
    \begin{subfigure}[t]{0.166666667\linewidth}
        \includegraphics[width=\linewidth]{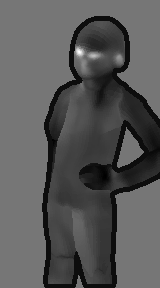}
        \caption{$n=2$}
    \end{subfigure}%%
    \begin{subfigure}[t]{0.166666667\linewidth}
        \includegraphics[width=\linewidth]{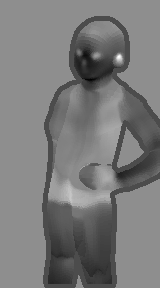}
        \caption{$n=4$}
    \end{subfigure}%%
    \begin{subfigure}[t]{0.166666667\linewidth}
        \includegraphics[width=\linewidth]{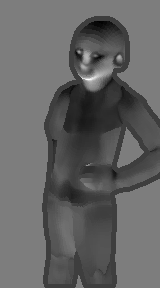}
        \caption{$n=6$}
    \end{subfigure}%%
    \begin{subfigure}[t]{0.166666667\linewidth}
        \includegraphics[width=\linewidth]{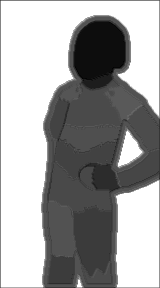}
        \caption{SPADE}
    \end{subfigure}
    \caption{
        Visualization of the norm of $\gamma$ for \modulationMetodShort where $f_\omega$ has $n\;$ layers (a-d), and (e) show SPADE  \cite{Park_2019} with 26 semantic regions.
        Note that \modulationMetodShort learns much more fine-grained details (\eg zoom in on head or fingers) than its semantic counterpart \cite{Park_2019}.}
    \label{fig:modulation_params}
\end{figure}
The global mapping network ($f_\omega$) adapts the smooth surface embedding into semantically meaningful surface-adaptive styles, which are not necessarily smooth.
For instance, this enables the generator to learn part-wise continuous styles with clearly defined semantic borders (\eg between two pieces of clothing).
We observe that a deeper mapping network learns higher-fidelity styles (\Cref{fig:modulation_params}), which improves image quality (shown in \Cref{sec:experimental_mapping_depth}).

Unlike prior semantic-based modulation \cite{Park_2019,tan2021INADE,Tan_2021CLADE}, \modulationMetodShort uses a denser and more informative representation that excels at human synthesis.
Semantic-based modulation learns spatially-invariant (but semantic-variant) styles \cite{Tan_2021CLADE}, which is reflected in \Cref{fig:modulation_params}.
These spatially-invariant parameters are efficient for natural image synthesis but translate poorly to the highly fine-grained task of human figure synthesis.
In contrast, \modulationMetodShort learns semantically detailed styles independent of pre-defined semantic regions.
%Learning this representation through semantic-based modulation is difficult, as it requires pre-defined semantic regions, and complicates simple operations (\eg downsampling of border regions).

\subsection{Variational Surface Adaptive Modulation}
\label{sec:variational_surface_adaptive_modulation}
A key limitation to \modulationMetodShort is that the appearance of the synthesized body depends on its spatial position.
Typically, an image-to-image generator inputs a latent code ($z$) directly to a 2D feature map through concatenation or additive noise.
However, this entangles the latent code with the spatial feature map, making the appearance of the generated person dependent on the position in the image.

Instead of inputting $z$ to a 2D feature map, we extend \modulationMetodShort to condition the mapping network on $z$; $\omega_i = f_\omega(e_i, z)$.
Now, $f_\omega$ transforms the latent variable $z$ to a surface-adaptive intermediate latent space ($\omega$), which is modulated onto the spatial feature map.
This naive extension of \modulationMetodShort allows the generator to directly relate latent factors of variations (\eg color of the shirt) to specific positions on the body.
Note that the variational modulation of \modulationMetodNoiseShort is independent of the spatial position of the body in the image, as $\gamma_i^\ell$ is determined solely from $(z, e_i)$.
This enables \modulationMetodNoiseShort  to modulate the style of the body invariant to image rotation and translation,
improving the ability of the generator to synthesize the same person independent of its spatial position
\footnote{Note that rotational invariance is not retained in the generator, as the generator is not rotationally invariant itself. However, adapting \modulationMetodNoiseShort with StyleGAN3-R \cite{stylegan3} produces a surface-guided rotationally invariant generator.}.

Adaptive modulation is an established technique in the literature for unconditional \cite{Huang_2017,karras2019style} and conditional modulation \cite{Park_2019,zhao2021comodGAN}.
However, the design of \modulationMetodNoiseShort is more expressive than current methods, and the explicit adaption of latent variables to surface locations independent of spatial position is unique to \modulationMetodNoiseShort.
The naive design of V-SAM originates from the plain representation of CSE, where equally expressive modulation techniques based on other representations (\eg DensePose or semantic maps) require much more engineering effort.
For example, current variational semantic-based modulation \cite{tan2021INADE,zhu2020sean} does not directly translate to human synthesis
\footnote{\Eg adapting \cite{tan2021INADE,zhu2020sean} for body parts requires class-specific latent variables that have to semantically match between related regions} and the styles generated by \modulationMetodNoiseShort are of higher fidelity.
Furthermore, the expressiveness of \modulationMetodNoiseShort significantly improves quality and disentanglement compared to previous methods, which we experimentally validate in \cref{sec:vsam_vs_other_modulation}.

\subsection{Discriminator Surface Supervision}
\label{sec:surface_regression}
Supervising the discriminator by teaching it to predict conditional information (instead of inputting it), is known to improve image quality and training stability \cite{odena2017conditional,schonfeld2020oasis}.
We propose a similar objective for surface embeddings.

We formulate the surface embedding prediction as a regression task.
We extend the discriminator with an FPN-head that outputs a continuous embedding for each pixel; $\hat{e}_i$.
Along with the adversarial objective, the discriminator optimizes a masked version of the smooth $L_1$ loss \cite{Girshick_2015};
\begin{equation}
    \mathcal{L}_{CSE}(\hat{e}, e) =  \sum_{i \in {h,w}} (1-M_i) \odot \text{smooth}_{L_1}(\hat{e}_i, e_i).
\end{equation}
Similarly, the generator objective is extended with the regression loss with respect to the generated image.
Unlike the original CSE loss \cite{neverova2020cse}, our objective is simpler as we assume a fixed embedding $e$ which is learned in advance.

Discriminator surface supervision explicitly encourages the discriminator to learn pixel-to-surface correspondences.
This yields a discriminator that provides highly detailed gradient signals to the generator, which considerably improves image quality.
In comparison to semantic-based supervision \cite{schonfeld2020oasis}, surface-supervision provides higher-fidelity feedback without relying on pre-defined semantic regions.
Finally, we found that additionally predicting "real" and "fake" areas (as in OASIS \cite{schonfeld2020oasis}) negatively affects training stability and that a FPN-head is more stable to train compared to a U-Net \cite{ronneberger2015u} architecture (as used in \cite{schonfeld2020oasis}).

\subsection{The Anonymization Pipeline}
Our proposed anonymization framework consists of two stages.
Initially, a CSE-based \cite{neverova2020cse} detector computes the location of humans, including a dense 2D-3D correspondence between the 2D image and a fixed 3D human surface.
Given the detected human, we zero-out pixels covering the human body and complete the partial image with a generative model.
Note that the masks generated from CSE \cite{neverova2020cse} do not cover areas that are "outside" of the human body, thus we dilate the mask to ensure that it covers clothing and hair.
We extend \cref{eq:f_w} with an additional pixel-independent learned parameter for the dilated regions (similar to $\omega_M$), to ensure a smooth transition between known areas and unknown dilated areas (without a surface embedding).

\begin{figure}
    \centering
    \begin{subfigure}[t]{0.166666667\linewidth}
    \includegraphics[width=\linewidth, trim={0 1.68em 0 0}, clip]{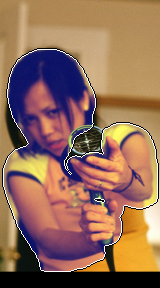}
    \includegraphics[width=\linewidth, trim={0 7em 0 4em}, clip]{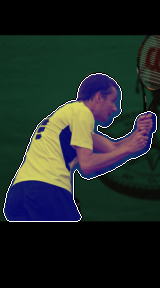}
    \includegraphics[width=\linewidth, trim={0 4em 0 2em}, clip]{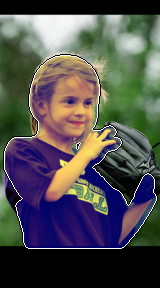}
    \includegraphics[width=\linewidth]{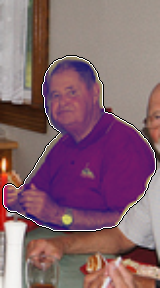}
    \includegraphics[width=\linewidth, trim={0 3.1em 0 3em}, clip]{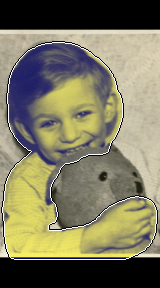}
    \captionsetup{justification=centering,textfont=tiny}
    \caption*{\tiny{Original}}
    \end{subfigure}%%
    \begin{subfigure}[t]{0.166666667\linewidth}
    \includegraphics[width=\linewidth, trim={0 1.68em 0 0}, clip]{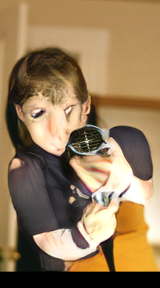}
    \includegraphics[width=\linewidth,trim={0 7em 0 4em}, clip]{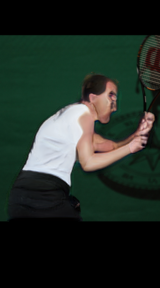}
    \includegraphics[width=\linewidth, trim={0 4em 0 2em}, clip]{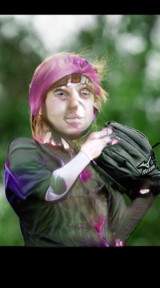}
    \includegraphics[width=\linewidth]{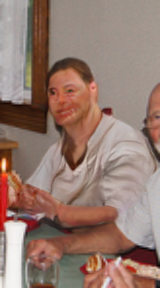}
    \includegraphics[width=\linewidth, trim={0 3.1em 0 3em}, clip]{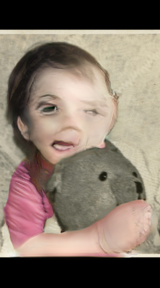}
    \captionsetup{justification=centering,textfont=tiny}
    \caption*{\tiny{SPADE}}
    \end{subfigure}%%
    \begin{subfigure}[t]{0.166666667\linewidth}
    \includegraphics[width=\linewidth, trim={0 1.68em 0 0}, clip]{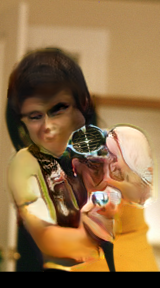}
    \includegraphics[width=\linewidth,trim={0 7em 0 4em}, clip]{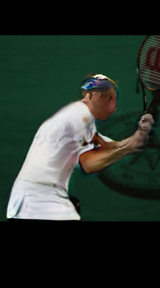}
    \includegraphics[width=\linewidth, trim={0 4em 0 2em}, clip]{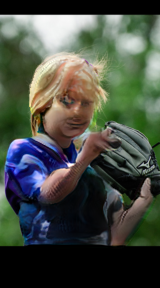}
    \includegraphics[width=\linewidth]{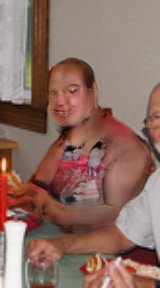}
    \includegraphics[width=\linewidth, trim={0 3.1em 0 3em}, clip]{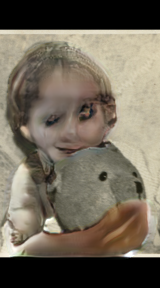}
    \captionsetup{justification=centering,textfont=tiny}
    \caption*{B}
    \end{subfigure}%%
    \begin{subfigure}[t]{0.166666667\linewidth}
    \includegraphics[width=\linewidth, trim={0 1.68em 0 0}, clip]{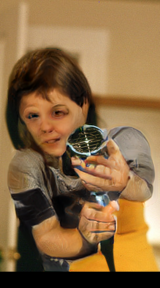}
    \includegraphics[width=\linewidth,trim={0 7em 0 4em}, clip]{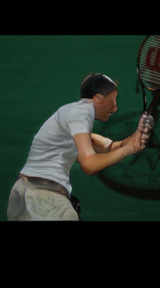}
    \includegraphics[width=\linewidth, trim={0 4em 0 2em}, clip]{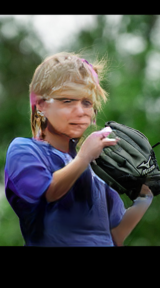}
    \includegraphics[width=\linewidth]{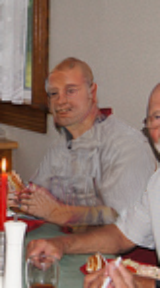}
    \includegraphics[width=\linewidth, trim={0 3.1em 0 3em}, clip]{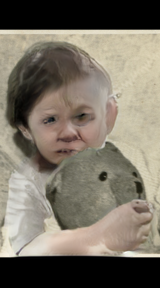}
    \captionsetup{justification=centering,textfont=tiny}
    \caption*{\tiny{D,n=0}}
    \end{subfigure}%%
    \begin{subfigure}[t]{0.166666667\linewidth}
    \includegraphics[width=\linewidth, trim={0 1.68em 0 0}, clip]{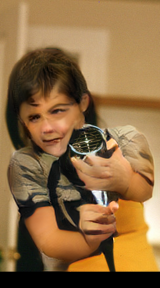}
    \includegraphics[width=\linewidth,trim={0 7em 0 4em}, clip]{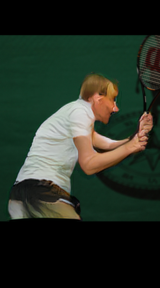}
    \includegraphics[width=\linewidth, trim={0 4em 0 2em}, clip]{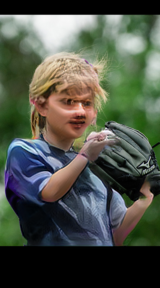}
    \includegraphics[width=\linewidth]{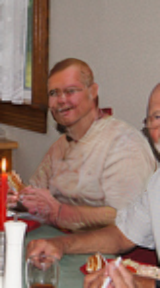}
    \includegraphics[width=\linewidth, trim={0 3.1em 0 3em}, clip]{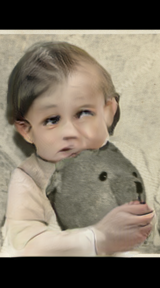}
    \captionsetup{justification=centering,textfont=tiny}
    \caption*{\tiny{D}}
    \end{subfigure}%%
    \begin{subfigure}[t]{0.166666667\linewidth}
    \includegraphics[width=\linewidth, trim={0 1.68em 0 0}, clip]{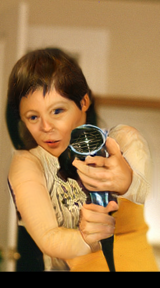}
    \includegraphics[width=\linewidth, trim={0 7em 0 4em}, clip]{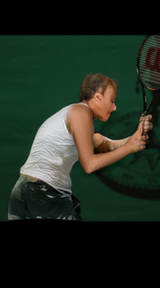}
    \includegraphics[width=\linewidth, trim={0 4em 0 2em}, clip]{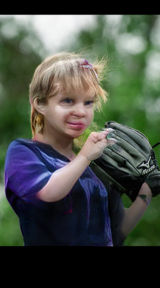}
    \includegraphics[width=\linewidth]{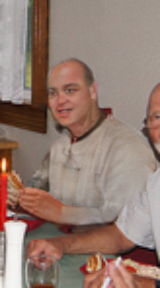}
    \includegraphics[width=\linewidth, trim={0 3.1em 0 3em}, clip]{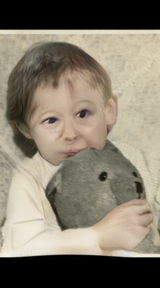}
    \captionsetup{justification=centering,textfont=tiny}
    \caption*{\tiny{E}}
    \end{subfigure}%%
    \caption{Synthesized images for the different model iterations in \Cref{tab:method_ablation}. \appendixQualitative includes random examples.}
    \label{fig:iterative_development}
\end{figure}

\begin{figure}
\centering
\begin{subfigure}[t]{0.2\linewidth}
    \includegraphics[width=\linewidth,trim={0 6em 0 1em}, clip]{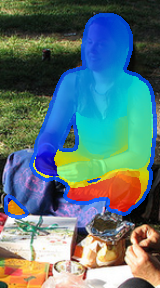}
    \includegraphics[width=\linewidth,trim={0 5em 0 2em}, clip]{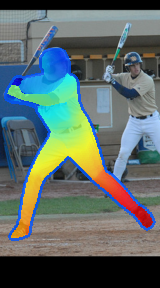}
    \includegraphics[width=\linewidth]{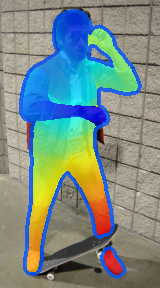}
    \includegraphics[width=\linewidth,trim={0 1em 0 .3em},clip]{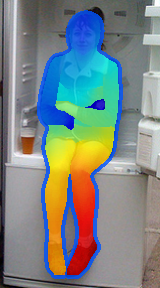}
\captionsetup{justification=centering,textfont=scriptsize}
\caption*{(a)}
\end{subfigure}%%
\begin{subfigure}[t]{0.2\linewidth}
    \includegraphics[width=\linewidth,trim={0 6em 0 1em}, clip]{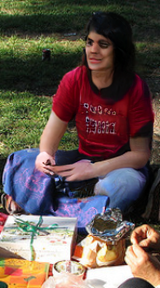}
    \includegraphics[width=\linewidth,trim={0 5em 0 2em}, clip]{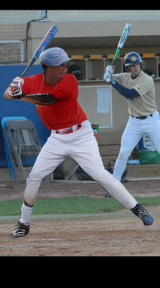}
    \includegraphics[width=\linewidth]{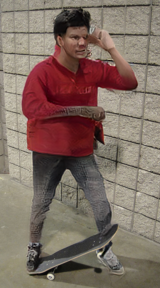}
    \includegraphics[width=\linewidth,trim={0 1em 0 .3em},clip]{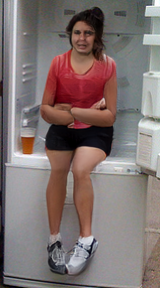}
    \captionsetup{justification=centering,textfont=scriptsize}
    \caption*{(b)}
\end{subfigure}%%
\begin{subfigure}[t]{0.2\linewidth}
    \includegraphics[width=\linewidth,trim={0 6em 0 1em}, clip]{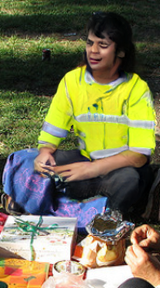}
    \includegraphics[width=\linewidth,trim={0 5em 0 2em}, clip]{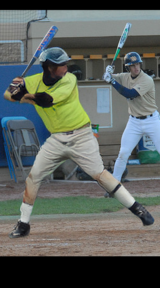}
    \includegraphics[width=\linewidth]{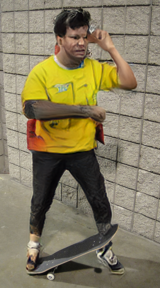}
    \includegraphics[width=\linewidth,trim={0 1em 0 .3em},clip]{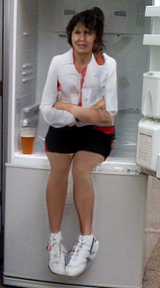}
    \captionsetup{justification=centering,textfont=scriptsize}
    \caption*{(c)}
\end{subfigure}%%
\begin{subfigure}[t]{0.2\linewidth}
    \includegraphics[width=\linewidth,trim={0 6em 0 1em}, clip]{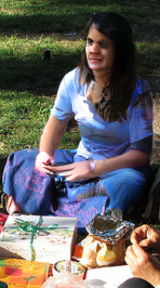}
    \includegraphics[width=\linewidth,trim={0 5em 0 2em}, clip]{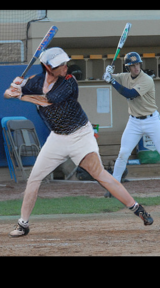}
    \includegraphics[width=\linewidth]{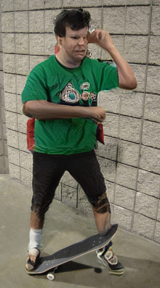}
    \includegraphics[width=\linewidth,trim={0 1em 0 .3em},clip]{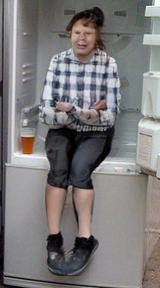}
    \captionsetup{justification=centering,textfont=scriptsize}
    \caption*{(d)}
\end{subfigure}%%
\begin{subfigure}[t]{0.2\linewidth}
    \includegraphics[width=\linewidth,trim={0 6em 0 1em}, clip]{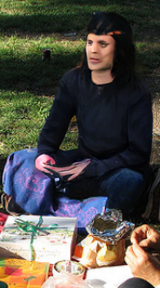}
    \includegraphics[width=\linewidth,trim={0 5em 0 2em}, clip]{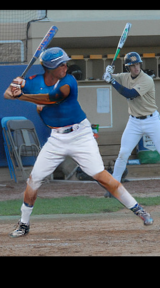}
    \includegraphics[width=\linewidth]{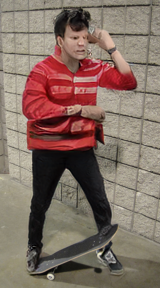}
    \includegraphics[width=\linewidth,trim={0 1em 0 .3em},clip]{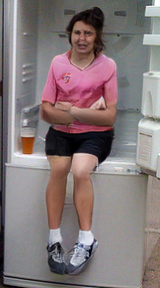}
    \captionsetup{justification=centering,textfont=scriptsize}
    \caption*{(e)}
\end{subfigure}%%
\caption{
    Config E diverse synthesis.
    (a) is the input, (b) is the generated image with truncation (t=0), and (c-e) are without truncation.
    \appendixQualitative includes random examples.}
\label{fig:cse_diverse}
\end{figure}

\section{Experiments}
We validate our design choices in \Cref{sec:surface_GANS_ablations} and compare \modulationMetodNoiseShort to alternative methods in \Cref{sec:vsam_vs_other_modulation}.
\Cref{sec:deep_fashion} ablate on the DeepFashion \cite{liu16DeepFashion} dataset for scene-independent human synthesis.
Finally \Cref{sec:anonymization_effect} evaluates the impact of anonymization for future computer vision development.
\appendixQuantitative and D include further evaluation.

\paragraph{Architecture Details}
We follow the implementation of StyleGAN2 \cite{karras2019analyzing} for our training setup.
The generator is a U-Net \cite{ronneberger2015u}, previously adapted for image-to-image translation \cite{Isola_2017}, and the discriminator is similar to the one of StyleGAN2.
The generator uses instance normalization for each convolution, operating only on standard deviation (\ie the mean is not used for normalization). 
The latent variable ($z$) is linearly projected and concatenated to the input of the decoder of the generator, unless it is inputted through modulation.
The baseline discriminator and generator has 8.5M and 7.4M parameters, respectively.
We use the non-saturating adversarial loss \cite{goodfellow2014generative} with epsilon penalty \cite{karras2018progressive} and r1-regularization \cite{mescheder2018training}.
We mask the r1-regularization by $M$, similar to  \cite{ConextualAttention2018Yu,hukkelaas2020image}.
Data augmentation is used for COCO-Body, including geometrical transforms, and color transforms.
Otherwise, we keep the training setup simple, with no feature matching loss \cite{Wang_2018}, or path length regularization \cite{karras2019analyzing}.
We set the dimensionality of $\omega_i$ and the fully-connected layers in $f_\omega\;$ to 512, and use 6 layers in $f_\omega$ unless stated otherwise.
\appendixDetails includes further details.

\paragraph{Dataset Details}
We validate our method on two datasets; a derived version of the COCO-dataset \cite{lin2014microsoft} (named \emph{COCO-Body}) for full-body anonymization and DeepFashion \cite{liu16DeepFashion} for static scene synthesis.
We will open-source the CSE-annotations for both datasets.

\begin{itemize}
    \item \emph{COCO-Body} contains cropped images from COCO \cite{lin2014microsoft}, where a single human is in the center of each image.
    Each image has automatically annotated CSE embeddings and a boolean mask indicating the area to be replaced.
    Note that each mask is dilated from the original CSE-embedding such that the mask covers all parts of the body.
    The dataset contains 43,053 training images and 10,777 validation images, with a resolution of $288 \times 160$.
    See \appendixAnonymization for more details.
    \item \emph{DeepFashion-CSE} includes images from the In-shop Clothes Retrieval Benchmark of DeepFashion \cite{liu16DeepFashion}, where we have annotated each image with a CSE embedding.
    It has 40,625 training images and 10,275 validation images, where each image is downsampled to $384 \times 256$.
    The dataset includes some errors in annotations, as no annotation validation is done.
\end{itemize}

\paragraph{Evaluation Details}
We follow typical evaluation practices for generative modeling.
We report Fr\'echet Inception Distance (FID) \cite{FID2017heusel}, Learned Perceptual Image Patch Similarity (LPIPS) \cite{zhang2018perceptual}, LPIPS Diversity \cite{zhu2017toward}, and Perceptual Path Length (PPL) \cite{karras2019style}.
FID, LPIPS, and LPIPS diversity is found by generating 6 images per validation sample, where the reported LPIPS is the average.
In addition, we report the face quality by evaluating FID for the face region (see \appendixDetails).
\appendixQuantitative includes all metrics for each model.

\subsection{Attributes of Surface-Guided GANs}
\label{sec:surface_GANS_ablations}
We iteratively develop the baseline architecture to introduce surface guidance.
\Cref{tab:method_ablation} (and \Cref{fig:iterative_development}) reflects that the addition of discriminator surface-supervision (config B) and surface modulation (config C/D) drastically improves image quality.
Note that adaptive modulation is only applied for the convolutional layers in the decoder.
Config E increases the model size of the generator and discriminator to 33M and 34M parameters, respectively.
The final generator produces high-quality and diverse results (\Cref{fig:cse_diverse}).
In addition, the conditional intermediate latent space $\omega$ is amenable to similar techniques as the latent space of StyleGAN \cite{karras2019style}, \eg the truncation trick \cite{Brock2018} and latent interpolation (ablated in \appendixQuantitative ).
\Cref{fig:cse_diverse} includes generated images with latent truncation.

\begin{table}[t]
    \caption{
        Iterative addition of surface guidance to the baseline.
        * $\mathcal{L}_{CSE}$ is applied to G and D, where G receives CSE-information by concatenation with the image
        }
    \label{tab:method_ablation}
    \centering
    \begin{adjustbox}{max width=\linewidth}
    \begin{tabular}{|cl|c|c|c|c|}
        \hline
        & Method & LPIPS $\downarrow$  & FID $\downarrow$ & PPL $\downarrow$ & Diversity $\uparrow$\\
        \hline
        \hline
        \textbf{A}:  & Baseline  & 0.237 & 7.4   & 26.7  & 0.162 \\
        \hline
        \textbf{B}: & A + $\mathcal{L}_{\text{CSE}}$* & 0.220 & 5.8   & 19.0  & 0.140 \\
        \hline
        \textbf{C}: &B + SAM    & 0.219 & 5.6   & 19.2  & 0.143 \\
        \hline
        \textbf{D}: &B + V-SAM  & 0.220 & 5.2   & \textbf{13.7}  & \textbf{0.166} \\
        \hline
        \textbf{E}:  & D + Larger D/G & \textbf{0.211} & \textbf{4.8}   & 15.1  & 0.161 \\
        \hline
    \end{tabular}
    \end{adjustbox}
 
\end{table}

\begin{table}
    \caption{
        Config D with different number of layers ($n$) in the mapping network ($f_\omega$).
        All other experiments use 6 layers.
        }
    \label{tab:mapping_depth}
    \centering
    \begin{adjustbox}{max width=\linewidth}
    \begin{tabular}{|c|c|c|c|}
        \hline
        $f_\omega$ depth ($n$)  &Face FID $\downarrow$ & FID $\downarrow$ & PPL $\downarrow$\\
        \hline
        \hline
        0          & 7.7   & 5.4   & 24.9  \\
        \hline
        2          & 8.0   & 5.4   & 19.7   \\
        \hline
        4          & 7.9   & 5.5   & 19.8  \\
        \hline
        6          & \textbf{7.4}   & \textbf{5.2}   & \textbf{13.7}  \\
        \hline
    \end{tabular}
    \end{adjustbox}    
\end{table}

\paragraph{Mapping Network Depth}
\label{sec:experimental_mapping_depth}

A deeper mapping network allows the generator to learn finer-grained modulation parameters, which we find to significantly improve image quality  and latent disentanglement (\Cref{tab:mapping_depth}).
Qualitatively, we observe that this significantly improves the quality of fine-grained regions (\eg the face and fingers, see \Cref{fig:iterative_development}).
We quantitatively validate this improvement through the FID of the upsampled face region (Face FID).

Furthermore, a deeper mapping network allows the generator to better disentangle the latent space
\footnote{Following Karras \etal \cite{karras2019style}, "disentangled latent space" refers to that the latent factors of variations are separated into linear subspaces.}, which is reflected by PPL.
The improved disentanglement is rooted in two design choices; first, \modulationMetodShort explicitly disentangles the variations of pose into surface-adaptive modulation.
Secondly, \modulationMetodNoiseShort allows the generator to easier control specific areas of the human body disentangled of the spatial image, by "unwarping" the fixed distribution $z$ to the surface-conditioned distribution $\omega$.

\paragraph{Affine Invariance Studies}
\modulationMetodNoiseShort is invariant to affine image-plane transformations, and thus, improves the ability of the generator to disentangle the latent representation from such transforms.
We quantitatively evaluate this with Peak Signal-to-Noise Ratio (PSNR), following \cite{zhang2019making},
\begin{equation}
    \mathbb{E}_{I,M,E,t \sim T} \text{PSNR} [t(G(\bar{I}, E)), G(t(\bar{I}), t(E))],
\end{equation}
where $\bar{I} = I \odot M$, E is the CSE embedding, G is the generator, and  $T$ is the distribution of vertical/horizontal image shifts.
$T\;$ is limited to translate by a maximum $\frac{1}{8}$ of the image width/height.
Similarly, we evaluate rotational invariance (limited to $\pm 90^{\circ}$) and horizontal flip.

\modulationMetodNoiseShort significantly improves  the baseline \wrt invariance to affine transformations (\cref{tab:modulation_comparisons}), as \modulationMetodNoiseShort is invariant to such transformations.
In comparison, \modulationMetodShort achieves similar scores as the baseline.
The aspect of affine-invariance is important for realistic anonymization, as the detection can induce slight shifts across frames.

\begin{table}[t]
    \caption{
        Comparison of \modulationMetodNoiseShort to alternative adaptive normalization methods.
        All methods are applied on top of config B.
        }
    \label{tab:modulation_comparisons}
    \centering
    \begin{adjustbox}{max width=\linewidth}
    \begin{tabular}{|l|c|c|c|c|c|c|c|c|}
        \hhline{----~----}
         & \multicolumn{3}{c|}{Affine Transformation} & &  \multicolumn{4}{c|}{Quality}\\
        Method & Translation $\uparrow$ & Rotation $\uparrow$ & Hflip $\uparrow$&  & Diversity $\uparrow$ & FID $\downarrow$ & PPL $\downarrow$ & Face FID $\downarrow$ \\ 
        \hhline{====~====}
%        \hline
%        \hhline{----~----}
        Config B   & 23.1  & 20.3  & 21.7  && 0.140 & 5.8   & 19.0  & 9.1   \\
        \hhline{----~----}
        B + SPADE \cite{Park_2019} & 22.5  & 19.8  & 20.7  && 0.150 & 5.9   & 20.6  & 9.7   \\
        \hhline{----~----}
        B + INADE \cite{tan2021INADE}  & 24.1  & 20.2  & 20.9 & & 0.140 & 5.8   & 19.5  & 9.4   \\
        \hhline{----~----}
        B + CLADE \cite{Tan_2021CLADE}   & 22.9  & 20.1  & 21.3 & & 0.138 & 5.7   & 16.9  & 8.9   \\
        \hhline{----~----}
        B + StyleGAN \cite{karras2019analyzing} & 25.5  & 20.9  & 21.6 & & 0.155 & 5.7   & 48.2  & 9.4   \\
        \hhline{----~----}
        B + CoMod \cite{zhao2021comodGAN}  & 24.5    & 20.6    & 21.6    & & 0.154  & 5.5  & 17.5 & 8.0 \\
        \hhline{====~====}
        B + SAM     & 23.8  & 20.7  & 21.4  && 0.143 & 5.6   & 19.2  & \textbf{7.4}   \\
        \hhline{----~----}
        B + V-SAM  & \textbf{26.1}  & \textbf{21.4}  & \textbf{22.5}  && \textbf{0.166} & \textbf{5.2}   & \textbf{13.7}  & \textbf{7.4}   \\

        \hhline{----~----}
    \end{tabular}
\end{adjustbox}
\end{table}

\paragraph{Computational Complexity}
\modulationMetodNoiseShort consists of two stages, the mapping network and layer-wise linear transformations.
Each layer-wise transformation is efficiently implemented as $1 \times 1\;$ convolution.
The mapping network is a sequence of fully-connected layers, which can be implemented as $1 \times 1\;$ convolution by using the spatial embedding map $e_i\;$ for each pixel i.
However, in practice, we find the nearest vertex embedding $e_k$ for each embedding $e_i$, and transform the 27K vertex-embeddings to $w_k$.
This results in a mapping network independent of image resolution.

\subsection{The Expressiveness of \modulationMetodNoiseShort}
\label{sec:vsam_vs_other_modulation}
We now analyze the expressiveness of \modulationMetodNoiseShort compared to well-established modulation techniques.
Specifically, we compare against adaptive instance normalization from StyleGAN2 \cite{karras2019analyzing}, co-modulation (CoMod) \cite{zhao2021comodGAN}, and variational \cite{tan2021INADE}/non-variational \cite{Park_2019,Tan_2021CLADE} semantic-based methods.
All methods are applied on top of Config B.

\cref{tab:modulation_comparisons} shows that \modulationMetodNoiseShort significantly improves upon previous modulation methods.
\modulationMetodNoiseShort generates higher-fidelity styles than both semantic-based modulation \cite{karras2019analyzing,zhao2021comodGAN} and spatially invariant modulation \cite{Park_2019,tan2021INADE,Tan_2021CLADE}, yielding a substantial improvement in image quality (FID).
This is especially prominent in semantically complex areas of the body (Face FID).
Note that the improvement of \modulationMetodNoiseShort over co-modulation \cite{zhao2021comodGAN} is significant, as it is approximately the same as increasing the number of parameters by 20M (config E vs D, \cref{tab:method_ablation}).
Furthermore, \modulationMetodNoiseShort improves latent disentanglement (PPL), originating from the expressive and explicit design of \modulationMetodNoiseShort.
Finally, \modulationMetodNoiseShort is more invariant to affine image-plane transformations.

\subsection{Synthesis of Humans in Static Scenes}
\begin{figure}[t]
    \centering
    \begin{subfigure}[t]{0.14285714286\linewidth}
    \includegraphics[width=\linewidth]{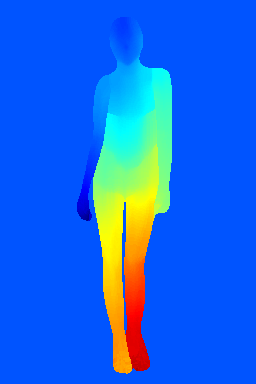}
    \includegraphics[width=\linewidth]{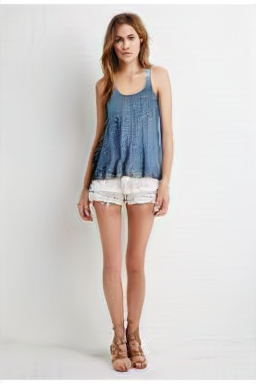}
    \includegraphics[width=\linewidth]{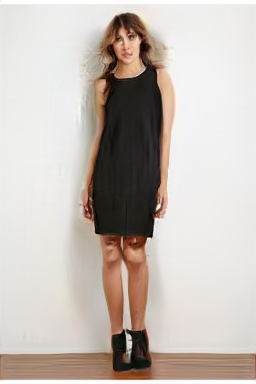}
    \includegraphics[width=\linewidth]{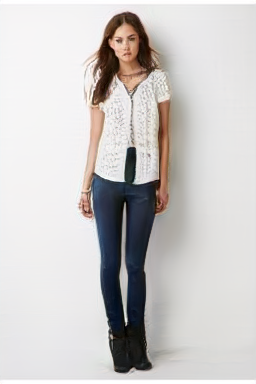}
    \includegraphics[width=\linewidth]{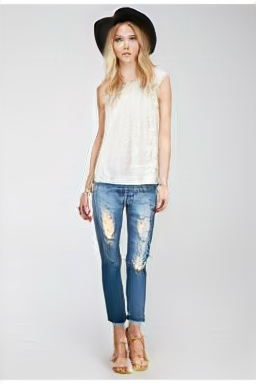}
    \end{subfigure}%%
    \begin{subfigure}[t]{0.14285714286\linewidth}
        \includegraphics[width=\linewidth]{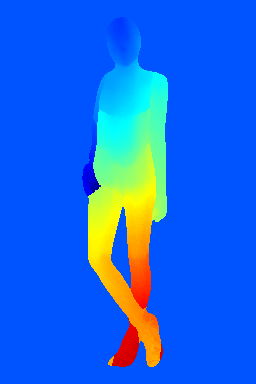}
        \includegraphics[width=\linewidth]{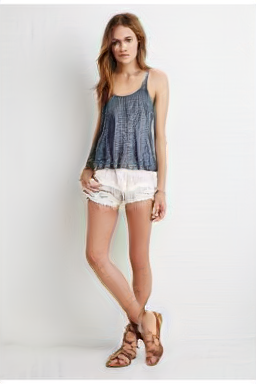}
        \includegraphics[width=\linewidth]{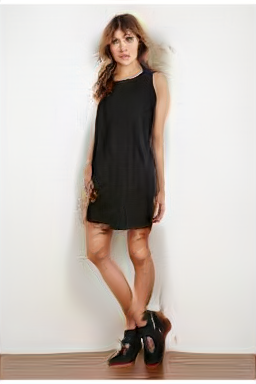}
        \includegraphics[width=\linewidth]{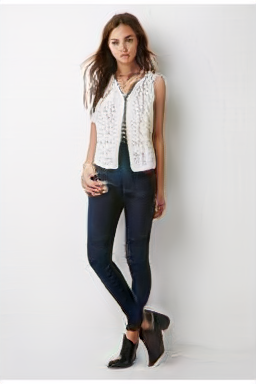}
        \includegraphics[width=\linewidth]{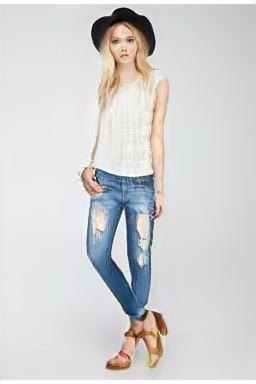}
    \end{subfigure}%%
    \begin{subfigure}[t]{0.14285714286\linewidth}
        \includegraphics[width=\linewidth]{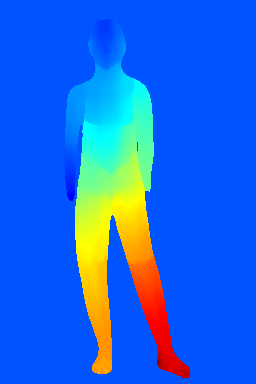}
        \includegraphics[width=\linewidth]{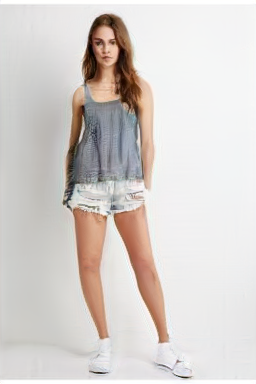}
        \includegraphics[width=\linewidth]{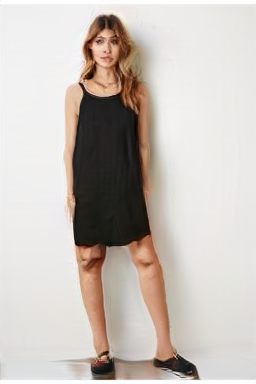}
        \includegraphics[width=\linewidth]{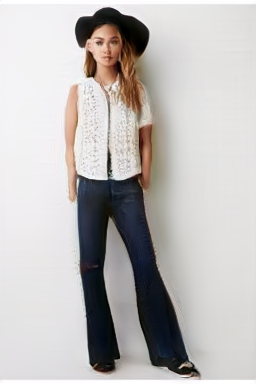}
        \includegraphics[width=\linewidth]{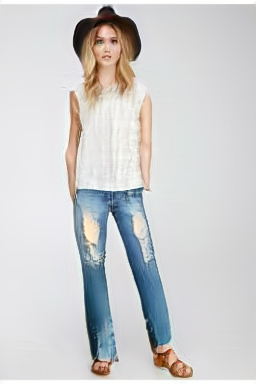}
    \end{subfigure}%%
    \begin{subfigure}[t]{0.14285714286\linewidth}
        \includegraphics[width=\linewidth]{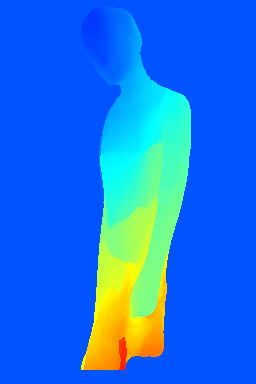}
        \includegraphics[width=\linewidth]{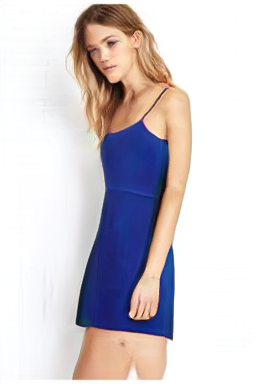}
        \includegraphics[width=\linewidth]{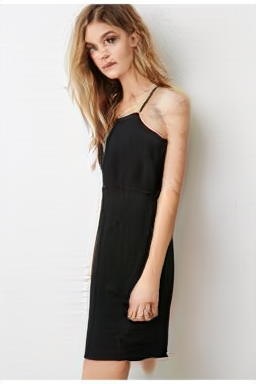}
        \includegraphics[width=\linewidth]{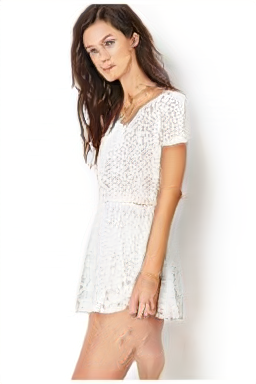}
        \includegraphics[width=\linewidth]{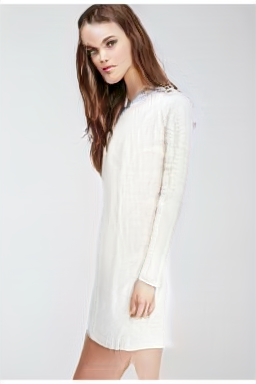}
    \end{subfigure}%%
    \begin{subfigure}[t]{0.14285714286\linewidth}
        \includegraphics[width=\linewidth]{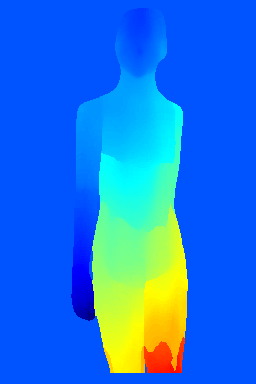}
        \includegraphics[width=\linewidth]{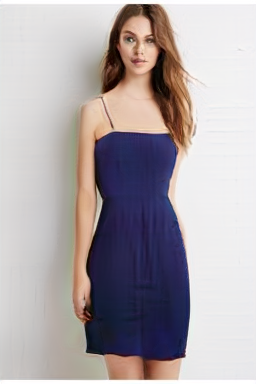}
        \includegraphics[width=\linewidth]{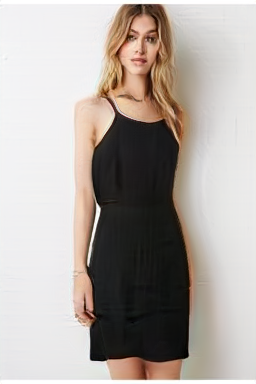}
        \includegraphics[width=\linewidth]{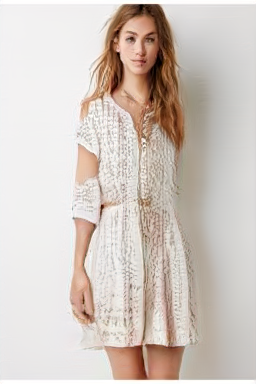}
        \includegraphics[width=\linewidth]{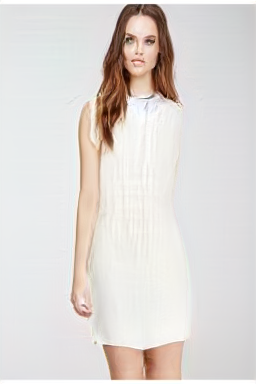}
    \end{subfigure}%%
    \begin{subfigure}[t]{0.14285714286\linewidth}
        \includegraphics[width=\linewidth]{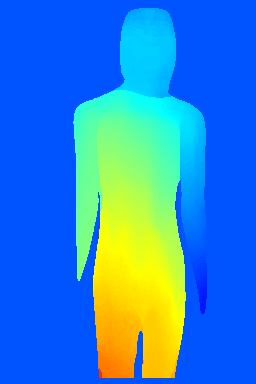}
        \includegraphics[width=\linewidth]{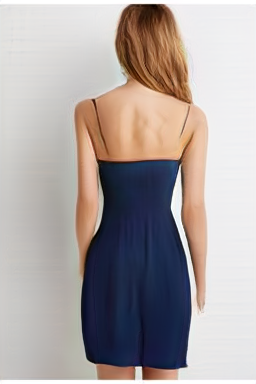}
        \includegraphics[width=\linewidth]{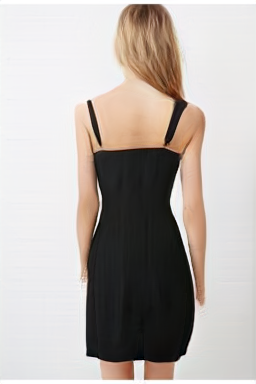}
        \includegraphics[width=\linewidth]{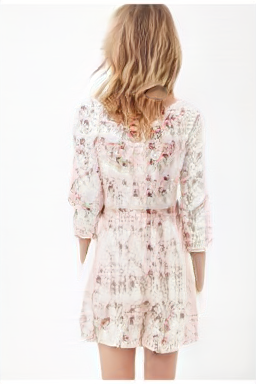}
        \includegraphics[width=\linewidth]{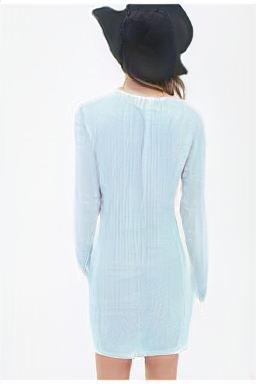}
    \end{subfigure}%%
    \begin{subfigure}[t]{0.14285714286\linewidth}
        \includegraphics[width=\linewidth]{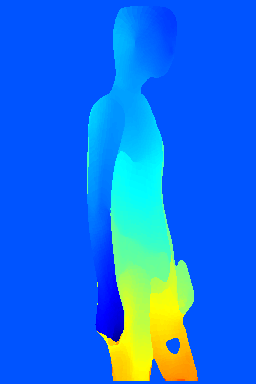}
        \includegraphics[width=\linewidth]{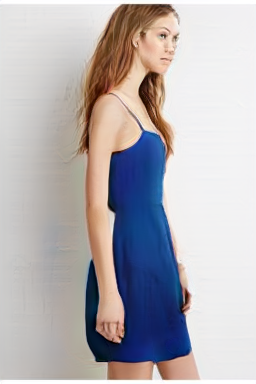}
        \includegraphics[width=\linewidth]{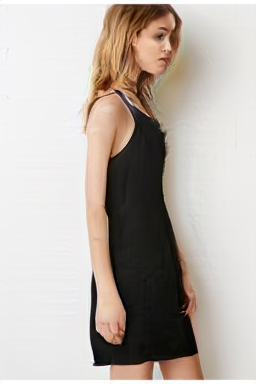}
        \includegraphics[width=\linewidth]{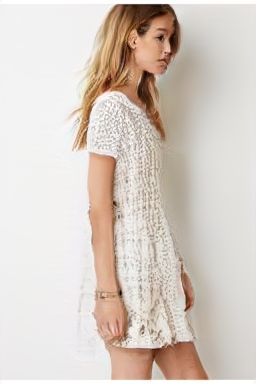}
        \includegraphics[width=\linewidth]{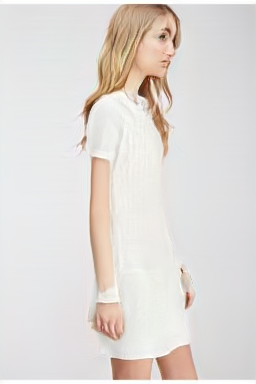}
    \end{subfigure}%%
    \caption{
        \modulationMetodNoiseShort can transfer attributes between poses by simply sampling the same latent variable $z$.
        Each row shows synthesized images with the same latent variable, but different input pose.}
    \label{fig:deep_fashion}
\end{figure}
\label{sec:deep_fashion}
We demonstrate that \modulationMetodNoiseShort excels at human synthesis for the DeepFashion \cite{liu16DeepFashion} dataset.
Following the design of SPADE \cite{Park_2019}, we design a decoder-only generator that synthesizes humans independent of any background image.

The disentangled and spatially-invariant latent space of \modulationMetodNoiseShort allows the generator to transfer attributes between poses.
By sampling the same latent variable $z$ for different poses, \modulationMetodNoiseShort is able to perform pose/motion transfer of synthesized humans (\Cref{fig:deep_fashion}) without any task-specific modeling choices (\eg including a texture encoder \cite{zhang2021pise}).
However, \modulationMetodNoiseShort is variant to 3D affine transformations that are not parallel to the imaging plane (\eg changing the depth of the scene).
This is reflected in \Cref{fig:deep_fashion}, where changing the depth of the scene significantly changes the synthesized person.
We believe that combining \modulationMetodNoiseShort with task-specific modeling choices from the pose/motion transfer literature \cite{men2020controllable,zhang2021pise} could resolve these issues.

\subsection{Effect of Anonymization for Computer Vision}
\begin{figure*}
    \centering
    \begin{subfigure}[]{0.2\linewidth}
        \includegraphics[width=.99\linewidth, trim={0 0 0 0}, clip]{figures/anonymization_comparison/000000507037.jpg}
        \captionsetup{justification=centering,textfont=scriptsize}
        \caption*{Original}
    \end{subfigure}%
    \begin{subfigure}[]{0.2\linewidth}
        \includegraphics[width=.99\linewidth, trim={0 0 0 0}, clip]{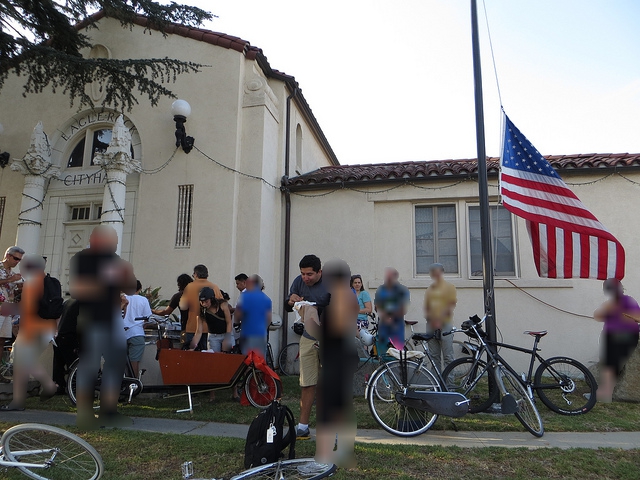}
        \captionsetup{justification=centering,textfont=scriptsize}
        \caption*{Pixelation $8 \times 8$}
    \end{subfigure}%
    \begin{subfigure}[]{0.2\linewidth}
        \includegraphics[width=.99\linewidth, trim={0 0 0 0}, clip]{figures/anonymization_comparison/000000507037_pixelation16.jpg}
        \captionsetup{justification=centering,textfont=scriptsize}
        \caption*{Pixelation $16 \times 16$}
    \end{subfigure}%
    \begin{subfigure}[]{0.2\linewidth}
        \includegraphics[width=.99\linewidth, trim={0 0 0 0}, clip]{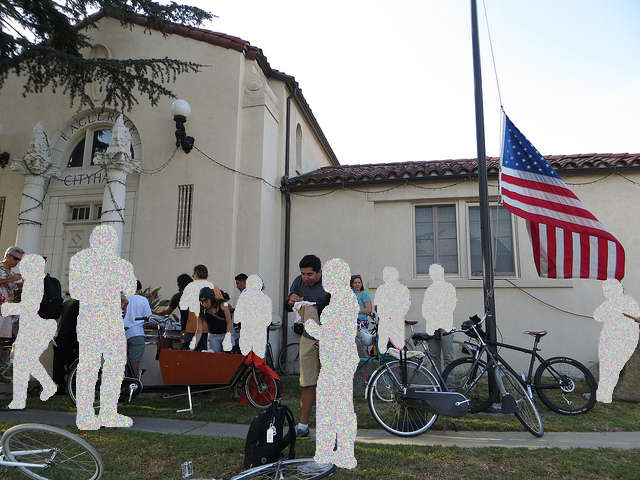}
        \captionsetup{justification=centering,textfont=scriptsize}
        \caption*{Masked out}
    \end{subfigure}%
    \begin{subfigure}[]{0.2\linewidth}
        \includegraphics[width=.99\linewidth, trim={0 0 0 0}, clip]{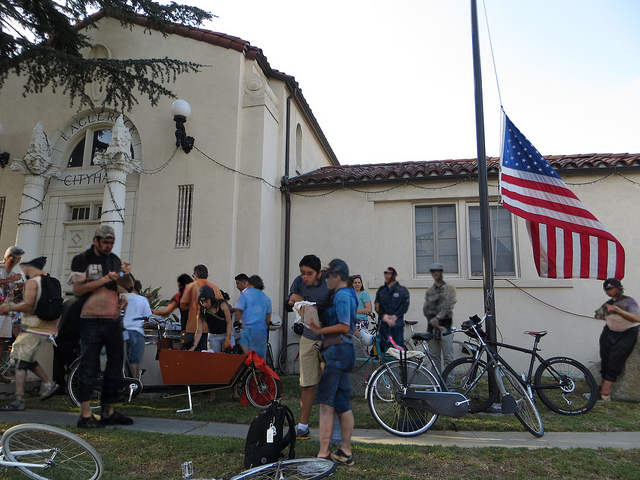}
        \captionsetup{justification=centering,textfont=scriptsize}
        \caption*{Ours}
    \end{subfigure}
    \caption{Different anonymization methods for an image from COCO \cite{lin2014microsoft} val2017. \appendixQualitative includes random examples.}
    \label{fig:anonymzation_qualitative_comparison}
\end{figure*}

\begin{table}[t]
    \caption{
        Instance segmentation mask AP on the COCO validation set \cite{lin2014microsoft}.
        The results are from a pre-trained Mask R-CNN \cite{he2017mask} \href{https://github.com/facebookresearch/detectron2/blob/335b19830e4ea5c5a74a085a04ff4a2f1a1dbf71/configs/COCO-InstanceSegmentation/mask_rcnn_R_50_FPN_3x.yaml}{R50-FPN-3x} from detectron2 \cite{wu2019detectron2} evaluated on different anonymized datasets.
        }
    \label{tab:coco_validation}
    \centering
    \begin{adjustbox}{max width=\linewidth}
    \begin{tabular}{|l|c|c|c|c|c|c|c|}
        \hline
        Validation Dataset & AP$_{50:95}$ $\uparrow$ & AP$_{50}$ $\uparrow$ & AP$_{75}$ $\uparrow$ & AP$_{\text{s}}$ $\uparrow$ & AP$_{\text{m}}$ $\uparrow$& AP$_{\text{l}}$ $\uparrow$ & AP$_\text{Person}$ $\uparrow$ \\
        \hline
        \hline
        Original                       & 37.2                           & 58.6                           & 39.9                           & 18.6                           & 39.5                           & 53.3                           & 47.7                           \\
        \hline
        \hline
        Mask Out                        & 32.8                           & 52.0                           & 35.1                           & 16.3                           & 34.6                           & 47.3                           & 27.5                           \\
        \hline
        $8 \times 8$ Pixelation        & 32.8                           & 51.8                           & 35.2                           & 16.4                           & 34.6                           & 47.2                           & 33.3                           \\
        \hline
        $16 \times 16$ Pixelation      & 33.4                           & 53.0                           & 35.7                           & 16.7                           & 35.0                           & 48.1                           & 38.4                           \\
        \hline
        Ours           & \textbf{34.6}                           & \textbf{55.0}                           & \textbf{37.0}                           & \textbf{17.1}                           & \textbf{36.8}                           & \textbf{50.0}                          & \textbf{44.9}                           \\
        \hline
    \end{tabular}
    \end{adjustbox}
\end{table}

\paragraph{Data Usability}
We analyze the effect of anonymization for future computer vision development by evaluating a pre-trained Mask R-CNN \cite{he2017mask} on the COCO dataset (results on PASCAL VOC \cite{everingham10pascal} are included in \appendixAnonymization ).
We anonymize all individuals that are detected by a pre-trained CSE-detector \cite{neverova2020cse}, where we use all detections with a confidence score higher than $0.1$.
We compare our framework to traditional anonymization methods (\autoref{fig:anonymzation_qualitative_comparison}).

Our method significantly improves $\text{AP}_\text{person}$ compared to traditional anonymization (\Cref{tab:coco_validation}), even pixelation, which is known to be questionable for anonymization \cite{gross2006model,neustaedter2006blur}.
However, we observe a notable drop in average precision for other object classes, which originates from two sources of error.
First, full-body anonymization removes objects that often appear with human figures.
For example, the "tie" class drops from 31\% AP to 1\% and "toothbrush" drops from 14.6\% to 6.2\%.
Secondly, the detections include false positives, yielding highly corrupted images when anonymizing these.
For example, the "zebra" class drops from 56.2\% to 48.0\%.
We observe insignificant degradation for objects that are rarely detected as person (\eg car, train, elephant).
Finally, surface-guided anonymization improves over traditional techniques for training purposes, which we validated on the anonymized COCO dataset (\appendixAnonymization ).
\label{sec:anonymization_effect}

\begin{table}[t]
    \centering 
    \caption{
        Re-identification mAP and rank-1 accuracy on Market1501 \cite{Zheng2015} using the official code of OSNet \cite{Zhou2019}.
        }
    \label{tab:market1501_re_identification}
    \begin{tabular}{|l|c|c|}
    \hline
        Anonymization & R1 $\downarrow$ & mAP $\downarrow$\\ 
        \hline 
        \hline 
        Original & 94.4 & 82.5 \\ 
        \hline
        Pixelation $8 \times 8$ & 67.8 & 39.6  \\
        Pixelation $16 \times 16$ & 86.6 & 66.4 \\
        Mask-out &  28.2 & 10.4 \\
        Face Anonymization \cite{hukkelaas2019DeepPrivacy} & 82.1 & 50.7 \\
        \hline
        Ours & \textbf{31.1} & \textbf{14.4} \\
        \hline
    \end{tabular}
    
\end{table}

%\noindent
\textbf{Anonymization Quality}
\Cref{tab:market1501_re_identification} evaluates the effect of anonymization for person re-identificaiton on the Market1501 \cite{Zheng2015} dataset.
Surface-guided GANs provide similar anonymization guarantees as masking out the region.
Meanwhile, face anonymization and pixelation yields a much higher re-identification rate, reflecting its worse anonymization guarantee.

\section{Conclusion}
We present a novel full-body anonymization framework that generates close-to-photorealistic and diverse humans in varying and complex scenes.
Our experiments show that guiding adversarial nets with dense pixel-to-surface correspondences strongly improves synthesis of high-fidelity textures for varying poses and scenes.
Finally, we demonstrate that our anonymization framework better retains data usability for future computer vision development compared to traditional anonymization.

\paragraph{Limitations}
Our contributions significantly improve the usability of anonymized data and generate new identities independent of the original.
However, our method has limitations that can compromise the privacy of individuals.
As with any anonymization method, our method relies on detection that is far from perfect \footnote{Current CSE-based detectors (R-101-FPN-DL-s1x \cite{wu2019detectron2}) has an average recall rate of 96.65\% (AR50) for human segmentation on COCO-DensePose \cite{Guler_2018}. Note that COCO-DensePose contains primarily high-resolution human figures.} and vulnerable to adversarial attacks.
Detection is improving every year and defense against adversarial attacks is currently  a large focus in the community \cite{kurakin2018adversarial}.
We believe that potential errors in detection can be circumvented with face detection as a fallback.

With the assumption of perfect detections, identification is still possible through gait recognition (when anonymizing videos), or through identity leaks in the CSE-embeddings.
We speculate that gait recognition can be mitigated by slightly randomizing the original pose between frames.
Furthermore, identity leaking through surface embeddings is possible, as they are regressed from the original image and could include identifying information.
We reduce this possibility by discretizing the regressed embedding into one of the 27K vertex-specific embeddings (\Cref{sec:method_generative}).

Surface-guided GANs significantly improve human figure synthesis for in-the-wild image anonymization.
Nevertheless, human synthesis is a complicated task, and many of the images generated by our method are recognizable as artificial by a human evaluator.
One of the limiting factors of our model is the dataset, where COCO-Body contains 40K images with a large variety.
This is relatively small compared to the 70K images in FFHQ \cite{karras2019style}, which is a considerably simpler task.
Our method applies data augmentation to mitigate this.
However, further extension with adaptive augmentation \cite{karras2020ada} or transfer learning could be fruitful.

\paragraph{Societal Impact}
We live in the age of Big Data, where personal information is the business model for many companies.
Recently introduced legislation has complicated data collection, requiring consent to store any data that contains personal information.
This can be a barrier to research and development, especially for the data-dependent field of computer vision.
We present a method that can better preserve the privacy of individuals, while retaining the usability of the data.
Nevertheless, our work focus on the synthesis of realistic humans, which has a potential for misuse.
The typical example is misuse of DeepFakes, where generative models can be used to create manipulated content with an intention to misinform.
Several solutions have been proposed, where the DeepFake Detection Challenge \cite{dolhansky2020deepfake} has increased the ability of models to detect manipulated content, and pre-emptive solutions such as model watermarking \cite{yu2021artificial} can mitigate the potential for misuse.
\paragraph{Acknowledgement}
The computations were performed on resources provided by the NTNU IDUN/EPIC computing cluster \cite{sjalander2019epic} and the Tensor-GPU project led by Prof. Anne C. Elster through support from The Department of Computer Science and The Faculty of Information Technology and Electrical Engineering, NTNU.
Furthermore, Rudolf Mester acknowledges the support obtained from DNV GL.
{\small
\bibliographystyle{ieee_fullname}
\bibliography{bib_files/main,bib_files/image_inpainting,bib_files/partial_convolution,bib_files/gan,bib_files/video_inpainting,bib_files/image_inpainting_trad,bib_files/probabilistic,bib_files/normalization,bib_files/multi_modal_image_to_image,bib_files/evaluation,bib_files/basic,bib_files/generative_models,bib_files/anonymization,bib_files/human_synthesis,bib_files/detection,bib_files/modulation,bib_files/smpl_detection}
}

\end{document}